\documentclass[letterpaper]{article} 
\usepackage{aaai25}  
\usepackage{times}  
\usepackage{helvet}  
\usepackage{courier}  
\usepackage[hyphens]{url}  
\usepackage{graphicx} 
\usepackage{subfigure}
\usepackage{natbib}  
\usepackage{caption} 
\usepackage{algorithm}
\usepackage{algorithmic}

\usepackage{amsmath}
\usepackage{amssymb}
\usepackage{multirow}
\usepackage{booktabs}
\usepackage{hwemoji}
\usepackage{float}

\usepackage{newfloat}
\usepackage{listings}
\DeclareCaptionStyle{ruled}{labelfont=normalfont,labelsep=colon,strut=off} 
\floatstyle{ruled}
\newfloat{listing}{tb}{lst}{}
\floatname{listing}{Listing}
\title{🍞 DAMPER: A Dual-Stage Medical Report Generation Framework with \\ Coarse-Grained MeSH Alignment and Fine-Grained Hypergraph Matching}
\author {
Xiaofei Huang\textsuperscript{\rm 1},
Wenting Chen\textsuperscript{\rm 2},
Jie Liu\textsuperscript{\rm 2},
Qisheng Lu\textsuperscript{\rm 1},
Xiaoling Luo\textsuperscript{\rm 1},
Linlin Shen\textsuperscript{\rm 1,3,4}
}
\affiliations {
\textsuperscript{\rm 1}Computer Vision Institute, College of Computer Science and Software Engineering, Shenzhen University, Shenzhen, China\\
\textsuperscript{\rm 2}Department of Electrical Engineering, City University of Hong Kong, Kowloon, Hong Kong\\
\textsuperscript{\rm 3}National Engineering Laboratory for Big Data System Computing Technology, Shenzhen University\\
\textsuperscript{\rm 4}Guangdong Provincial Key Laboratory of Intelligent Information Processing\\
2450101014@mails.szu.edu.cn,\{wentichen7-c, jilu.ee\}@my.cityu.edu.hk,qishenglu85@gmail.com,\{lluo,llshen\}@szu.edu.cn\\
}

\usepackage{bibentry}

\begin{document}

\maketitle

\begin{abstract}
	Medical report generation is crucial for clinical diagnosis and patient management, summarizing diagnoses and recommendations based on medical imaging. However, existing work often overlook the clinical pipeline involved in report writing, where physicians typically conduct an initial quick review followed by a detailed examination. Moreover, current alignment methods may lead to misaligned relationships. To address these issues, we propose DAMPER, a dual-stage framework for medical report generation that mimics the clinical pipeline of report writing in two stages. In the first stage, a MeSH-Guided Coarse-Grained Alignment (MCG) stage that aligns chest X-ray (CXR) image features with medical subject headings (MeSH) features to generate a rough keyphrase representation of the overall impression. In the second stage, a Hypergraph-Enhanced Fine-Grained Alignment (HFG) stage that constructs hypergraphs for image patches and report annotations, modeling high-order relationships within each modality and performing hypergraph matching to capture semantic correlations between image regions and textual phrases. Finally,the coarse-grained visual features, generated MeSH representations, and visual hypergraph features are fed into a report decoder to produce the final medical report. Extensive experiments on public datasets demonstrate the effectiveness of DAMPER in generating comprehensive and accurate medical reports, outperforming state-of-the-art methods across various evaluation metrics.
	
\end{abstract}

\section{Introduction}
Medical reports are essential for highlighting clinical findings and summarizing diagnoses and recommendations, which are crucial for clinical decision-making and patient management. However, the rise in medical imaging has made traditional manual report generation time-consuming and error-prone, potentially affecting patient safety. Therefore, automated medical report generation techniques have become increasingly important~\cite{MMTN}.

With the widespread application of deep learning in the medical field, advancements have also been made in various tasks such as medical visual question answering \cite{DALNet-WSE}, rectal cancer diagnosis \cite{jiansong}, clinical decision making \cite{liu2024medchain} and retinal disease detection \cite{luo2023mvcinn, luo2024lesion, xu2024hacdr}. Automatic algorithms for medical report generation have been widely studied. Some methods~\cite{cnn-rnn1,rg1} directly map input chest X-ray (CXR) images to medical reports using CNNs and RNNs, while others focus on effective image-text alignment to facilitate report generation. The latter methods believe that cross-modal disparity between images and texts may affect the mapping from CXR images to reports. They primarily rely on two types of alignment: coarse-grained and fine-grained alignment. Coarse-grained alignment methods~\cite{M2KT,chen2024fine,chen2024medical} align entire image features with whole report features, whereas fine-grained alignment methods~\cite{CMM,M2KT} aim to match specific image regions with corresponding descriptions in the report. However, these methods still face two main challenges when using alignment for report generation.

\begin{figure}[t]
	\centering
	\includegraphics[width=\columnwidth]{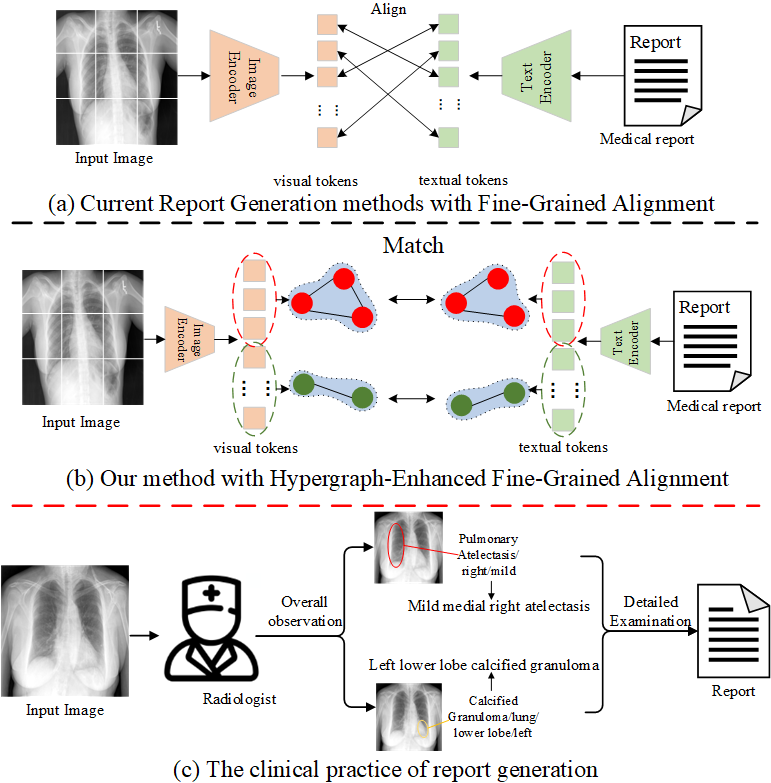}
	\caption{(a) Represents the previous research works on fine-grained alignment. (b) Our research work on fine-grained alignment in this paper. (c) The motivation for our approach: Mimicking the real process of physicians writing reports.}
	\label{example}
\end{figure}

The first challenge in automatic medical report generation is that current methods often generate reports directly from medical images, overlooking the clinical pipeline involved in report writing. Physicians typically conduct a quick review of images to form an overall impression, identifying major structures and abnormalities, followed by a detailed examination of specific regions for accurate diagnosis~\cite{jiedu}, as shown in Fig.~\ref{example} (c). While some approaches~\cite{mei2024phraseaug,jing2017automatic} consider the clinical practice of writing reports, they rely on clinical history or use disease labels to generate key phrases. However, these methods require additional effort to obtain extra patient information, which may not always be readily available due to privacy concerns surrounding medical data. Hence, it is imperative to consider the clinical practice of report generation to ensure the comprehension and accuracy of the generated reports. 

Another challenge in current fine-grained alignment methods is that they directly match image tokens with words, leading to misaligned relationships. In medical reports, detailed pathological findings are often described for different anatomical regions~\cite{tanida2023interactive}, as shown in Fig. \ref{example} (c), "left lower lobe calcified granuloma" (yellow region) and "mild right lung atelectasis" (red region). This implies that certain phrases in the report are correlated with the specific anatomical regions. Existing methods~\cite{MKMIA,UAR} achieve fine-grained alignment by establishing the relation between each image region and each word token one by one, as shown in Fig. \ref{example} (a). However, this direct token-to-word matching approach can potentially fragment semantic information across different patches or words, resulting in incomplete and unreasonable alignments between individual patches and words. Therefore, it is crucial to consider intra-modal relationships during the fine-grained alignment process in report generation.

To address these challenges, we propose the \textbf{D}u\textbf{a}l-Stage \textbf{M}edical Re\textbf{p}port Gen\textbf{er}ation Framework with Coarse-Grained MeSH Alignment and Fine-Grained Hypergraph Matching (DAMPER). This framework is inspired by the actual process of medical report writing and aims to incorporate clinical practices. DAMPER consists of a MeSH-guided Coarse-Grained Alignment (MCG) to mimic radiologists' overall observations and a Hypergraph-Enhanced Fine-Grained Alignment (HFG) to replicate detailed examinations. We found that MeSH information extracted from medical reports effectively captures key content and details, aiding in simulating the report writing process in clinical practice. \textbf{MCG stage} utilizes MeSH information to represent the patient's overall health status and align CXR images with MeSH. It consists of a MeSH Encoding module that transforms MeSH into graph embeddings, a GAN-Based MeSH-CXR Alignment (MCA) module that aligns multi-view CXR images with these embeddings, and a CXR-to-MeSH generation (CMG) module that maps the aligned CXR features into MeSH. The MCA module employs adversarial learning to ensure that the multi-view features generated are close to the graph embeddings. To further enhance overall observation, the CMG uses attention fusion and a MeSH Decoder to convert multi-view features into MeSH. The hypergraph structure effectively captures complex intra-modal relationships during fine-grained alignment. \textbf{ HFG stage}, we address potential misalignments between image patches and words by introducing hypergraphs to consider intra-modal relationships, preserving higher-order relationships throughout the alignment process. HFG includes Intra-Patient Report-CXR Alignment (Intra-RCA), Inter-Patient Report-CXR Alignment (Inter-RCA), and a report decoder. Intra-RCA constructs hypergraphs and performs node matching. We use hyperedges to connect patches within image regions and words within report phrases, capturing the complex relationships among modality-specific elements. Through node matching, we effectively align image patches with their corresponding report phrases. Inter-RCA employs contrastive learning for hypergraph embedding matching, optimizing feature representations to enhance accuracy in matching image regions with report phrases. After attention-based fusion of outputs from MCA, CMG, and Intra-RCA, the results are fed into the report decoder to generate a comprehensive medical report. To manage varying numbers of input images, we introduce a Bernoulli indicator in the generative adversarial and graph matching processes, ensuring model robustness even with missing visual information. Our contributions can be summarized as follows:
\begin{itemize}
	\item We introduce DAMPER, a medical report generation model designed with a dual-stage architecture that integrates coarse-grained and fine-grained alignments. This model emulates the process physicians follow when creating reports, enabling the generation of comprehensive and accurate medical documentation.
	\item To simulate the radiologist's holistic observation, we designed the MCG stage, which aligns CXR visual features with MeSH information.
	\item To simulate the detailed examination following the initial overall observation, we developed the HFG stage, aligning image regions with report phrases while preserving the high-order relationships among image patches.
	\item Experimental evaluations of DAMPER were conducted on two public datasets, IU-Xray and MIMIC-CXR. The results demonstrate that DAMPER outperforms state-of-the-art methods across various evaluation metrics.
\end{itemize}

\begin{figure*}[h]
	\centering
	\includegraphics[width=0.9\textwidth]{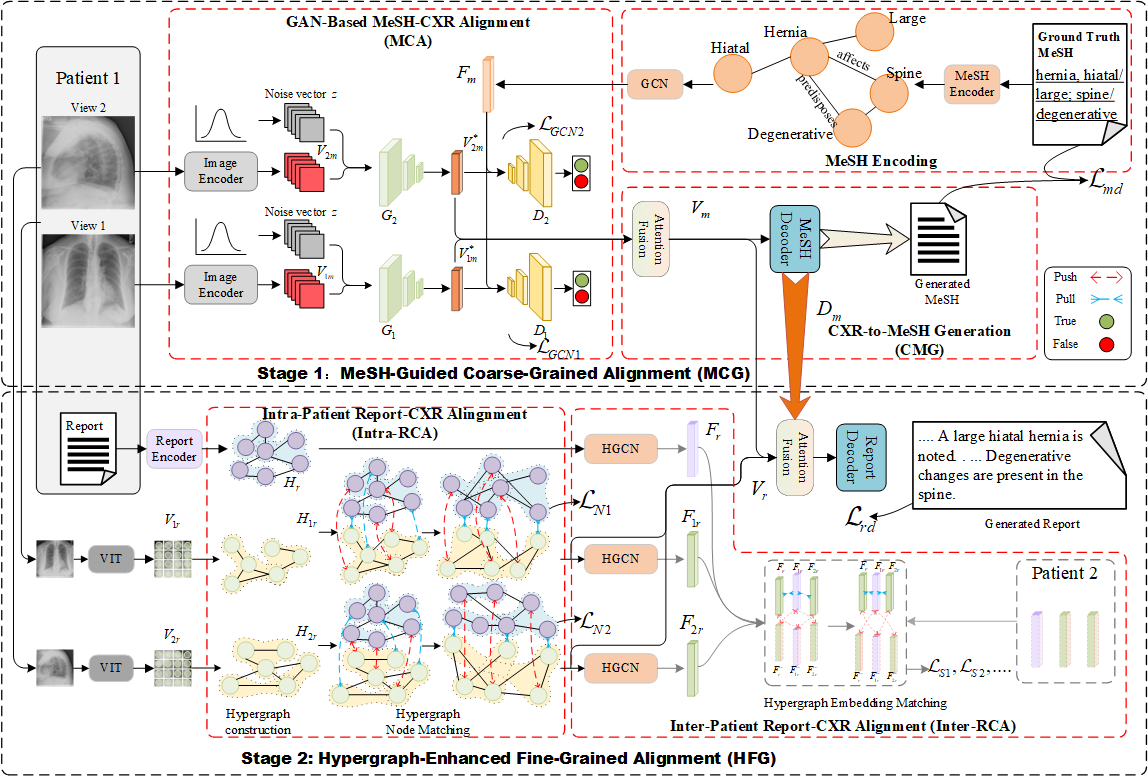}
	\caption{Overview of the proposed DAMPER framework. In the MCG stage, the MCA module extracts coarse-grained visual features aligned with MeSH terms, which are then input into the CMG. The HFG stage includes Intra-RCA and Inter-RCA modules that acquire detailed information corresponding to the report. This information, along with the output from MCA and CMG, is fed into the report decoder to generate the final medical report.}
	\label{DAMPER}
\end{figure*}

\section{Related Work}

\subsection{Cross-Modality Alignment}
Cross-modal alignment involves matching and integrating features from different modalities, such as images and text, to effectively combine visual and semantic information. Recent advancements in this field include techniques like CLIP, GAN, and graph matching.

The CLIP model excels in coarse-grained feature interaction through pretraining on large image-text datasets, demonstrating strong zero-shot learning capabilities \cite{clip}. CLIP's success has spurred further research, such as SoftCLIP \cite{SoftCLIP} and mCLIP \cite{mclip}, which have refined alignment techniques. However, since MeSH consists of keywords, it may not fully capture the semantic relationships between images, and CLIP-like models often require extensive training data for specialized domains. In contrast, GAN offers greater flexibility, making it more suitable for the coarse-grained alignment task in this context.

Graph matching techniques, such as GEM \cite{liu2024gem} and Bi-VLGM \cite{chen2024bi}, provide a novel approach to managing complex cross-modal relationships. These methods utilize graph structures to represent intricate associations between image regions and text, allowing for more precise alignment. Our study builds on this by exploring hypergraph matching for fine-grained alignment, which enhances the representation of complex relationships between visual features and text phrases.

\subsection{Medical Report Generation}

Medical report generation is a complex task that involves transforming medical images into accurate and coherent textual descriptions. To improve the quality of generated text, researchers introduced Transformer-based methods \cite{M2KT, KdTNet}, significantly enhancing the coherence and contextual understanding of generated reports. Subsequently, some researchers attempted to incorporate additional information into the decoding process to further improve the quality and accuracy of the generated reports \cite{rstnet,ppked}. 

Despite the significant advancements in report generation quality and medical accuracy achieved by these methods, they share a common limitation: the MeSH is not fully explored. MeSH terms are crucial elements that encapsulate disease conditions and health status, and they are essential for generating reports the accurately reflect pathological information. In light of this, we propose incorporating MeSH information into the report generation process to better simulate the real-world report writing workflow, with the aim of producing more accurate radiology reports.

\section{Methods}

\subsection{Overview}
DAMPER is a two-stage framework that simulates the medical report generation process in clinical practice. The first stage, MCG, includes a MeSH Encoding, MCA , and CMG modules. During the MCG stage, we extract MeSH graph embeddings $F_m$ and image representations $V_{im}$ using the MeSH Encoding and Image Encoder. Then, the MCA module employs a GAN to generate coarse-grained visual features $V^*_{im}$ aligned with $F_m$, which are input into the CMG module to produce phrase representations that capture the overall impression. The second stage, HFG, involves features extraction, Intra-RCA, Inter-RCA, and report generation. HFG constructs hypergraphs $H_r$ and $H_{ir}$ based on report phrases and image regions, achieving precise matching through loss functions $\mathcal{L}_N$ and $\mathcal{L}_S$ in Intra-RCA and Inter-RCA. Finally, the outputs $V_m$, $D_m$, and $V_r$ from MCA, CMG, and Intra-RCA are fused and passed into the report decoder to generate the textual report. The overall workflow is illustrated in Fig. \ref{DAMPER}.

\subsection{Stage 1: MeSH-Guided Coarse-Grained Alignment}
~\citet{jiedu} noted that in clinical report writing, physicians first form an overall impression before analyzing details. Existing methods either require more patient information~\cite{mei2024phraseaug} or overlook underlying pathologies when considering clinical practices~\cite{jing2017automatic}. To address this, we designed the MCG stage, which includes the MeSH Encoding, MCA, and CMG. The primary objective of MCG is to optimize the extraction of coarse-grained visual information and align it with MeSH, simulating the radiologist’s overall assessment to accurately capture pathological information and guide the subsequent report generation.

\subsubsection{MeSH Encoding} 
We use Clinical-BERT \cite{clinic-bert} to extract word information from MeSH as node features $N_m$. Reference to RGM \cite{rgm}, we obtain the corresponding edge relationship features $E_m$ to construct the MeSH graph $G_m=(N_m, E_m)$. This graph is then input into a GCN to obtain its embedding representation:
\begin{equation}
	F_{m} = GCN(G_m).
\end{equation}	

\subsubsection{GAN-Based MeSH-CXR Alignment (MCA)}
Given the flexibility of GANs and the need to obtain visual features corresponding to MeSH, we developed the MCA module. In practical applications, multiple medical images, denoted as $I_i$ for $i=1,2,...,H$, may need to be processed simultaneously. For each image $I_i$, we use ResNet to extract its overall visual feature $V_{im}$. During the adversarial learning, we use the generator of GAN to generate feature vectors $V^*_{im}$ that resembles the target vector $F_{m}$ based on the initial feature vector $V_{im}$. The loss for this adversarial generation process for view $i$ is expressed as:
\begin{equation}
	\resizebox{.9\hsize}{!}{$
	\begin{aligned}
		\mathcal{L}_{GANi} &=\mathbb{E}_{x\sim p_{data}(x)}[\log D_i(x|V_{im})]+\mathbb{E}_{z\sim p_z(z)}[\log(1\\&-D_i(G_i(z,V_{im})|V_{im}))]
		+\|F_{m}-G_i(z,V_{im})\|^2.
	\end{aligned}$}
\end{equation}

The loss function consists of three components: the discriminator's loss on real data, the discriminator's loss on generated data, and the mean squared error between the generated features and the MeSH features. Here, $D_i$ and $G_i$ represent the discriminator and generator in the GAN for view $i$, respectively, while $z$ is the noise vector generated from a normal distribution.

\subsubsection{CXR-to-MeSH Generation (CMG)}
To enhance the holistic representation and provide more accurate feature representations for report generation, we designed the CMG module. It employs an attention mechanism to fuse the multimodal information $V^*_{-m}$, obtaining a comprehensive representation from all viewpoints, can be expressed as:
\begin{gather}
		V_m=PA(Q,K,V)=Softmax(\frac{QK^T}{\sqrt{d_k}})V,\\
		Q=XW_Q,K=XW_K,V=XW_V,\\
		X=[\delta^{M}\cdot V^*_{1m};\delta^{M}\cdot V^*_{2m};...;\delta^{M}\cdot V^*_{Hm}],
\end{gather}where $W_Q, W_K$, and $W_V$ are learnable parameter matrices used to generate queries, keys, and values, respectively. $\delta\in\{0,1\}$ is a Bernoulli indicator introduced to enhance the model's robustness in scenarios where certain views are missing. In this manner, we obtain the coarse-grained radiological representation $V_m$ after the MCA process. The $V_m$ is fed into the MeSH decoder to generate the corresponding MeSH representation, which we train using a negative log-likelihood loss function:
\begin{equation}
	\resizebox{.6\hsize}{!}{$
	\mathcal{L}_{md}=-\sum_{t=1}^{T}\log P(y_t|y_{<t},V_m),$}
\end{equation}
here, $y_t$ represents the $t$-th word in the target sequence, and $y_{<t}$ represents the preceding words. To further distill the essential MeSH information, we apply a SoftPool operation on the final layer's hidden features, obtaining a refined MeSH feature representation. 

The loss function for the entire MCG stage can be expressed as follows: 	
\begin{equation}
	\resizebox{.6\hsize}{!}{$
	\mathcal{L}_{MCG}=\sum_{i=1}^{H}\delta^{M}\mathcal{L}_{GANi}+\mathcal{L}_{md}.$}
\end{equation}

\subsection{Stage 2: Hypergraph-Enhanced Fine-Grained Alignment}
Previous fine-grained alignment methods directly matched image regions with words~\cite{MKMIA,UAR}, often neglecting higher-order semantic relationships within each modality. To achieve precise alignment while preserving complex intra-modal connections, we designed the HFG stage, which includes: Features Extraction, Intra-RCA and Inter-RCA, and  Report Generation. The HFG stage aims to refine the extracted fine-grained visual information to match detailed medical expressions in the report, simulating the thorough examination process in clinical report writing.

\subsubsection{Features Extraction}
We utilize Clinical-BERT to extract textual information from medical report: $R={\mathrm Clinical-BERT(report)}$. For a given medical image $I_i$, we employ a Vision Transformer (VIT) model to extract its fine-grained features: $V_{ir}=VIT(I_i)$.	

\subsubsection{Intra-Patient Report-CXR Alignment (Intra-RCA)}
To capture intra-modal relationships and align image regions with report phrases, we designed the Intra-RCA module. We use RGM and KNN clustering to build the report hypergraph $H_r=(N_r,E_r)$, which is input into a HGCN to obtain the overall relational representation $F_{r} = HGCN(H_r)$. Based on $V_{ir}$, we construct a visual feature hypergraph $H_{ir}=(N_{ir},E_{ir})$ for each view $i$ and process it using a HGCN to obtain the overall representation $F_{ir}=HGCN(H_{ir})$.

We employ node feature matching to reduce the distance between fine-grained visual features and similar textual expressions while increasing the disparity with dissimilar expressions. The node matching loss is defined as follows:
\begin{equation}
	\resizebox{.9\hsize}{!}{$
		\begin{aligned}
			\mathcal{L}_{Ni}(N_r,N_{ir})=\max[\tau-m_{node}(N_r,N_{ir})+m_{node}(N_r^{-},N_{ir}),0]\\
			+ \max[\tau-m_{node}(N_r,N_{ir})+m_{node}(N_r, N_{ir}^{-}),0],
		\end{aligned}$}
\end{equation}
where $\tau$ is the margin parameter, $N_r^{-}$ and $ N_{ir}^{-}$ denote the negative samples in $N_r$ and $N_{ir}$, respectively. The graph node matching function $m_{node}(\cdot)$ is used to compute the sum of the maximum similarity values for each node match, which is specifically calculated as follows:
\begin{equation}
	\resizebox{.9\hsize}{!}{$m_{node}(N_{I},N_{T})=\sum_{0\leq i\leq m,0\leq j\leq l}\max[cos(N_{Ii},N_{Tj}),0],$}
\end{equation}
where $cos(\cdot)$ denotes the cosine similarity function. 

\begin{table*}[t]
	\centering
	\caption{Performance of DAMPER and Related Models on IU-Xray and MIMIC-CXR Test Sets}
	\label{tab1}
	\begin{center}
		\resizebox{\textwidth}{!}{
			\begin{tabular}{c| c c c c c c |c c c c c c}
				\toprule[1.25pt]
				~ & \multicolumn{6}{c|}{IU-Xray} & \multicolumn{6}{|c}{MIMICI-CXR}\\
				\hline
				Methods & B-1 & B-2 & B-3 & B-4 & M &R-L & B-1 & B-2 & B-3 & B-4 & M &R-L\\
				\hline
				AlignTransformer (MICCAI'21) &0.484 &0.313 &0.225 &0.173 &0.204 &0.379 &0.378 &0.235 &0.156 &0.112 &0.158 &0.283 \\
				R2Gen (EMNLP'20) &0.470 &0.304 &0.219 &0.165 &0.187 &0.371 &0.353 &0.218 &0.145 &0.103 &0.142 &0.277 \\
				R2GenCMN (ACL'21) &0.470 &0.304 &0.222 &0.170 &0.191 &0.358 &0.348 &0.206 &0.135 &0.094 &0.136 &0.266\\
				PPKED (CVPR'21)&0.483 &0.315 &0.224 &0.168 &0.190 &0.376 &0.360 &0.224 &0.149 &0.106 &0.149 &0.284\\
				KiUT (CVPR'23)&0.525 &0.360 &0.251 &0.185 &0.242 &0.409 &0.393 &0.243 &0.159 &0.113 &0.160 &0.285\\
				MMTN (AAAI'23) &0.486 &0.321 &0.232 &0.175 &- &0.375 &0.379 &0.238 &0.159 &0.116 &0.161 &0.283\\
				COMG (WACV'24)&\textbf{0.536} &0.378 &0.275 &0.206 &0.218 &0.383&0.363 &0.235 &0.167 &0.124 &0.128 &0.290\\
				MedM2G (CVPR'24) &0.533 &0.369 &0.278 &0.212 &- &\textbf{0.416} &\textbf{0.412} &0.260 &0.179 &0.142 &- &\textbf{0.309}\\
				\cline{1-13}
				\textbf{DAMPER}&0.520 &\textbf{0.383} &\textbf{0.300} &\textbf{0.225} &\textbf{0.284} &0.397&0.402 &\textbf{0.284} &\textbf{0.227} &\textbf{0.193} &\textbf{0.289} &0.301\\
				\bottomrule[1.25pt]
			\end{tabular}	
		}
	\end{center}
\end{table*}

\begin{table}[h]
	\centering
	\caption{Comparison of CE metrics between DAMPER and related models on the MIMIC-CXR dataset.}
	\label{tab-1}
	\begin{center}
		\resizebox{\columnwidth}{!}{
			\begin{tabular}{c | c c c}
				\toprule[1.25pt]
				Methods &Precision &Recall &F1\\
				\hline
				R2Gen (EMNLP'20) &0.333 &0.273 &0.276\\
				R2GenCMN (ACL'21) &0.334 &0.275 &0.278\\
				CvT2DistilGPT2 (AIM'23) &0.367 &0.418 &0.391\\
				KiUT (CVPR'23) &0.371 &0.318 &0.321\\
				M2KT (MIA'23) &0.420 &0.339 &0.352\\
				COMG (WACV'24) &0.424 &0.291 &0.345\\
				PromptMRG (AAAI'24) &0.501 &\textbf{0.509} &0.476\\
				\hline
				DAMPER &\textbf{0.512} &0.473 &\textbf{0.507}\\
				\bottomrule[1.25pt]
			\end{tabular}	
		}
	\end{center}
\end{table}

\subsubsection{Inter-Patient Report-CXR Alignment (Inter-RCA)} 
We designed the Inter-RCA module, which employs contrastive learning to achieve hypergraph embedding alignment between positive samples and the hardest negative samples, further enhancing the matching accuracy between image regions and report phrases. We utilize hypergraph embeddings $F_r$ and $F_{ir}$ to represent the features of the entire graphs $H_r$ and $H_{ir}$, respectively, encompassing both node features and relational characteristics. We introduce a hypergraph embedding matching loss to maximize the distance between negative and positive samples within a mini-batch. The loss function is defined as:
\begin{equation}
	\resizebox{.88\hsize}{!}{$
		\begin{aligned}
			\mathcal{L}_{Si}(F_r,F_{ir})=\max[\gamma-\cos(F_r,F_{ir})+\cos(F_r^-,F_{ir}),0]\\
			+\max[\gamma-\cos(F_r,F_{ir})+\cos(F_r,F_{ir}^-),0],
		\end{aligned}$}
\end{equation}
where $\gamma$ is the margin parameter. $F_r^-$ and $F_{ir}^-$ represent the most challenging negative samples for $F_r$ and $F_{ir}$, respectively, within the mini-batch. 

We concatenate the obtained $V_{ir}$ to derive the fine-grained feature information $V_r$ from all views. 

\subsubsection{Report Generation}
The report decoder generates a complete medical report by integrating $V_m$, $V_r$, and $D_m$ as inputs using an attention mechanism. By combining these multi-granularity features, it produces a comprehensive report that includes accurate medical subject headings and detailed descriptions. We train the report decoder using a negative log-likelihood loss:
\begin{equation}
	\resizebox{.7\hsize}{!}{$
	\mathcal{L}_{rd}=-\sum_{t=1}^{T}\log P(y_t|y_{<t},D_m, V_m, V_r). 
	$}
\end{equation}	
The loss in the HFG stage can be expressed as:
\begin{equation}
	\resizebox{.85\hsize}{!}{$
	\mathcal{L}_{HFG}=\sum_{i=1}^{H}\delta^{R}\cdot (\mathcal{L}_{Ni}(N_r,N_{ir})+\mathcal{L}_{Si}(F_r,F_{ir}))+\mathcal{L}_{rd}.$}
\end{equation}

The overall loss function of the DAMPER framework combines the losses from the MCG and HFG components, denoted as:
\begin{equation}
	\mathcal{L}_{total} = \mathcal{L}_{MCG}+\mathcal{L}_{HFG}.
\end{equation}

By leveraging this diversified loss function, DAMPER effectively transforms multi-granularity visual details into accurate and comprehensive medical reports.

\begin{figure*}[h]
	\centering
	\includegraphics[width=0.9\textwidth]{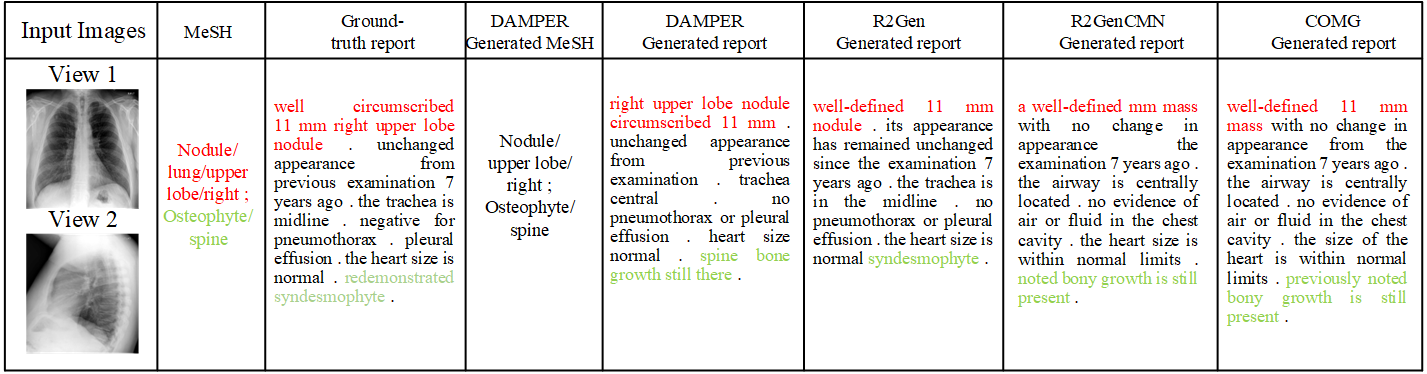}
	\caption{Visualization of report generation examples includes the input image in the first column, followed by the corresponding MeSH and report in the 2nd and 3rd columns. The 4th column shows the MeSH generated by DAMPER, while the 5th column backward shows reports from various models. MeSH are highlighted in red or green, with related sentences in the generated reports marked accordingly.}
	\label{dingx}
\end{figure*}

\section{Experiment}

\subsection{Experimental setup}

\subsubsection{Datasets} We conducted experiments on two publicly available datasets: IU-Xray \cite{IU-Xray}, MIMIC-CXR \cite{MIMIC}. \textbf{IU-Xray} comprises 7,470 pairs of CXR images and corresponding diagnostic reports, each including MeSH information. We used the same data preprocessing and splitting approach as R2Gen, with a training/validation/testing ratio of 7:1:2. \textbf{MIMIC-CXR} is the largest publicly available radiology dataset, containing 377,110 CXR images and 227,835 reports from 65,379 patients. We employed the official data split with a training/validation/testing ratio of 7:1:2. Notably, MIMIC-CXR does not include MeSH, so we used the CheXpert \cite{chexpert} labeler to generate MeSH based on the reports.

\subsubsection{Implementation Details} We conducted our experiments using the PyTorch framework on a single NVIDIA V100 GPU. Both the MeSH decoder and the report decoder utilize a 3-layer Transformer decoder architecture. The model was trained using the Adam optimizer with a batch size of 16 and a learning rate of 1e-4. 

We compare our method against the following models: AlignTransformer \cite{AlignTransformer}, R2Gen \cite{R2Gen}, R2GenCMN \cite{R2GenCMN}, PPKED \cite{ppked}, KiUT \cite{KiUT}, MMTN \cite{MMTN}, COMG \cite{COMG}, MedM2G \cite{MedM2G}, PromptMRG \cite{PromptMRG}, M2KT \cite{M2KT}, CvT2DistilGPT2 \cite{CvT2DistilGPT2}, RECAP \cite{RECAP}, ICON \cite{ICON} and ORGAN \cite{ORGAN}. These models represent a diverse range of architectures and methodologies in the field of medical report generation, spanning from deep learning-based encoder-decoder frameworks to multimodal fusion techniques, showcasing the variety and innovation in current research. Through comparisons with these state-of-the-art models, we aim to comprehensively evaluatethe performance and advantages of DAMPER in generating high-quality medical reports.

\subsubsection{Evaluation Metrics} In the evaluation process, we used metrics: BLEU (B) \cite{bleu}, METEOR (M) \cite{meteor}, and ROUGE-L (R-L) \cite{rouge}. The BLEU metric can reflect whether the generated reports use correct medical terminology and common expressions. METEOR better reflects the semantic accuracy of the generated reports, rather than just literal matching, helping to assess whether the reports accurately convey medical concepts. ROUGE-L aids in evaluating whether the reports describe various medical findings and diagnoses in the correct order. 

To further evaluate the model's clinical performance, we incorporated clinical efficacy (CE) metrics, including precision, recall, and F1-score, following the same approach as \cite{CvT2DistilGPT2} and \cite{M2KT}. On the MIMIC-CXR dataset, we utilized the rule-based CheXpert labeler to extract disease labels from both ground-truth and generated reports, enabling the calculation of CE metrics.

\begin{table*}[h]
	\centering
	\caption{Description of ablation conditions and experimental results on the IU-Xray dataset. A $\checkmark$ indicates inclusion.}
	\label{tab0}
	\begin{center}
		\resizebox{\textwidth}{!}{
			\begin{tabular}{c | c c c c | c c c c c c}
				\toprule[1.25pt]
				Status &MCA &CMG &Intra-RCA &Inter-RCA &B-1 &B-2 &B-3 &B-4 &M &R-L\\
				\hline
				w/o CMG+ Intra-RCA + Inter-RCA &$\checkmark$   &  &  & &0.467 &0.313 &0.210 &0.155 &0.188 &0.371\\
				w/o Intra-RCA + Inter-RCA &$\checkmark$  &$\checkmark$ &  &  &0.501 &0.347 &0.221 &0.163 &0.230 &0.382\\
				w/o Inter-RCA &$\checkmark$  &$\checkmark$ &$\checkmark$ &  &0.511 &0.358 &0.260 &0.198 &0.260 &0.368\\
				w/o Intra-RCA &$\checkmark$  &$\checkmark$ &  &$\checkmark$ &0.507 &0.350 &0.258 &0.188 &0.258 &0.353\\
				w/o CMG &$\checkmark$ & &$\checkmark$ &$\checkmark$ &0.497 &0.341 &0.216 &0.156 &0.193 &0.370\\
				w/o MCA &  &$\checkmark$ &$\checkmark$ &$\checkmark$ &0.478 &0.328 &0.214 &0.153 &0.186 &0.339\\
				w/o MCG &   &  &$\checkmark$ &$\checkmark$ &0.471 &0.328 &0.215 &0.152 &0.203 &0.348\\	
				DAMPER &$\checkmark$  &$\checkmark$ &$\checkmark$ &$\checkmark$ &\textbf{0.520} &\textbf{0.383} &\textbf{0.300} &\textbf{0.225} &\textbf{0.284} &\textbf{0.397}\\
				\bottomrule[1.25pt]
			\end{tabular}	
		}
	\end{center}
\end{table*}

\subsection{Comparison with state-of-the-art}

In the comparative analysis of the DAMPER model with state-of-the-art approaches on the IU-Xray and MIMIC-CXR datasets, Table \ref{tab1} highlights the significant advantages of DAMPER. Compared to other advanced models, DAMPER demonstrated superior performance across both datasets, achieving the highest scores particularly in BLEU-(2-4) and METEOR metrics. Although its performance on BLEU-1 and ROUGE-L was slightly lower than the best-performing models, the substantial improvement in METEOR is especially noteworthy. These results underscore the significant progress made by DAMPER in generating high-quality, semantically accurate medical reports.

Furthermore, Table \ref{tab-1} presents the CE comparison between DAMPER and related models on the MIMIC-CXR dataset, revealing that DAMPER outperformed previous approaches in terms of both precision and F1-score. This demonstrates DAMPER's ability to provide more reliable support for clinical report generation.

A qualitative comparison of DAMPER with open-source models such as R2Gen, R2GenCMN, and COMG is presented, focusing on the accuracy of MeSH terms in the generated reports, as illustrated in Fig. \ref{dingx}. While the reports may appear similar, DAMPER provides more precise descriptions of health conditions and pathological details critical for diagnosis and treatment. For instance, DAMPER accurately captures critical details such as "well circumscribed 11 mm right upper lobe nodule" and "redemonstrated syndesmophyte," whereas other models often give incomplete or inaccurate descriptions. This demonstrates DAMPER's advantage in integrating MeSH information to produce more comprehensive reports reflecting the patient's health status. More examples can be found in the appendix.

\subsection{Ablation Studies}

To validate the effectiveness and rationale of each component, we designed the following model configurations (see Table \ref{tab0}) and conducted ablation experiments on the IU-Xray dataset. The results are presented in Table \ref{tab0}.

\subsubsection{The effectiveness of MCA} 
Comparing the test results of w/o MCA with DAMPER reveals a significant performance decline on the IU-Xray dataset when the MCA module is removed, with the METEOR metric dropping by nearly 0.1. Additionally, when comparing w/o CMG with w/o MCG, w/o CMG outperforms w/o MCG in all metrics, with the most notable difference in the ROUGE-L. These results confirm that the MCA module, by focusing on the overall features of the images, significantly enhances the model's performance and the semantic quality of the generated reports.

\subsubsection{The effectiveness of CMG} 
When comparing the performance of DAMPER with w/o CMG, we observed a significant drop in performance after removing the CMG, with all metrics decreasing by at least 0.03 and METEOR dropping by 0.09. Further analysis of configurations w/o Intra-RCA + Inter-RCA and w/o MCA also showed a decline in performance across all metrics when CMG was removed, though the differences were less pronounced. These findings highlight the importance of CMG in medical report generation, as it enhances semantic accuracy and content consistency, improving the model's understanding of pathology and health status, and thereby the quality and accuracy of generated reports. Examples of the generated reports can be found in the appendix.

\subsubsection{The effectiveness of Intra-RCA} 
By comparing the results of DAMPER with w/o Intra-RCA and w/o Inter-RCA with w/o Intra-RCA + Inter-RCA, we observed a significant performance drop when Intra-RCA was removed. This decline is likely due to the difficulty in effectively aligning phrases in the report with visual features without the supervision of Intra-RCA. This finding highlights the critical role of Intra-RCA in generating accurate phrase-level text.

\subsubsection{The effectiveness of Inter-RCA} 
Comparing DAMPER with w/o Inter-RCA and w/o Intra-RCA with w/o Intra-RCA + Inter-RCA, we observed a decline in model performance. The decrease was more pronounced when Intra-RCA was removed, indicating that Intra-RCA is more critical to DAMPER's performance than Inter-RCA.

\begin{table}[h]
	\centering
	\caption{The effectiveness of Ground-truth MeSH information for report generation on the IU-Xray dataset.}
	\label{tab4}
	\begin{center}
		\resizebox{\columnwidth}{!}{
			\begin{tabular}{c|c c c c c c}
				\toprule[1.25pt]
				Methods & B-1 & B-2 & B-3 & B-4 & M &R-L\\
				\hline
				Disease Labels as MeSH &0.486 &0.307 &0.213 &0.157 &0.187 &0.365\\
				Ground-truth MeSH (DAMPER) &\textbf{0.520} &\textbf{0.383} &\textbf{0.300} &\textbf{0.225} &\textbf{0.284} &\textbf{0.39}7\\
				\bottomrule[1.25pt]
			\end{tabular}	
		}
	\end{center}

\end{table}

\subsubsection{The effectiveness of MeSH information}
To evaluate the importance of MeSH information in report generation, we replaced the original MeSH terms in the IU-Xray dataset with those generated by Chexpert, referred to as "Disease Labels as MeSH." The results, shown in Table \ref{tab4}, indicate that this replacement leads to a decrease in model performance. However, compared to the results in Table \ref{tab1}, where MCG was omitted, the performance with generated MeSH, while inferior to DAMPER, still surpasses the case where MeSH is entirely discarded. This suggests that although generated MeSH is less effective than the original, it still contributes positively to model performance.

\begin{table}[h]
	\centering
	\caption{Test Comparison Results for the Validation of Hypergraph effectiveness in DAMPER on the IU-Xray Dataset.}
	\label{tab5}
	\begin{center}
		\resizebox{\columnwidth}{!}{
			\begin{tabular}{c|c c c c c c}
				\toprule[1.25pt]
				Methods & B-1 & B-2 & B-3 & B-4 & M &R-L\\
				\hline
				w/o Hypergraph + w/ Graph  &0.492 &0.343 &0.220 &0.158 &0.204 &0.369\\
				DAMPER &\textbf{0.520} &\textbf{0.383} &\textbf{0.300} &\textbf{0.225} &\textbf{0.284} &\textbf{0.397}\\
				\bottomrule[1.25pt]
			\end{tabular}	
		}
	\end{center}
\end{table}

\subsubsection{The effectiveness of Hypergraph}
In this study, we utilized hypergraphs to establish correspondences between visual region features and report phrases. To assess the superiority of hypergraphs over traditional graphs in this context, we replaced the hypergraphs in DAMPER with graphs, defining this condition as w/o Hypergraph + w/ Graph. The comparative results are presented in Table \ref{tab5}.

The results indicate that replacing the original hypergraphs with graphs leads to a decline in model performance, particularly noticeable in the METEOR metric. This highlights the distinct advantage of hypergraphs in capturing medical terminology and effectively representing the complex relationships between visual features and text. Thus, hypergraphs are better suited for modeling the intricate associations between visual regions described in reports and the corresponding medical concepts.

\subsection{Visualization of Coarse-and-Fine-Grained Alignment}
We visualized the alignment examples of MCG and HFG in Fig. \ref{t-SNE} and Fig. \ref{align} to demonstrate the effectiveness of MCA and hypergraph matching.

\subsubsection{Visualization Coarse-Grained Alignment}
\begin{figure}[h]
	\centering
	\subfigure[MeSH and View 1]{%
		\includegraphics[width=0.45\columnwidth]{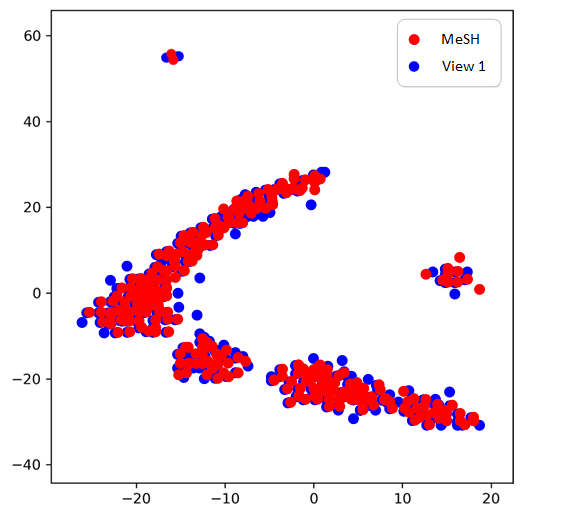} 
		\label{fig:imageA}
	}\hspace{0.1cm}
	\subfigure[MeSH and View 2]{%
		\includegraphics[width=0.45\columnwidth]{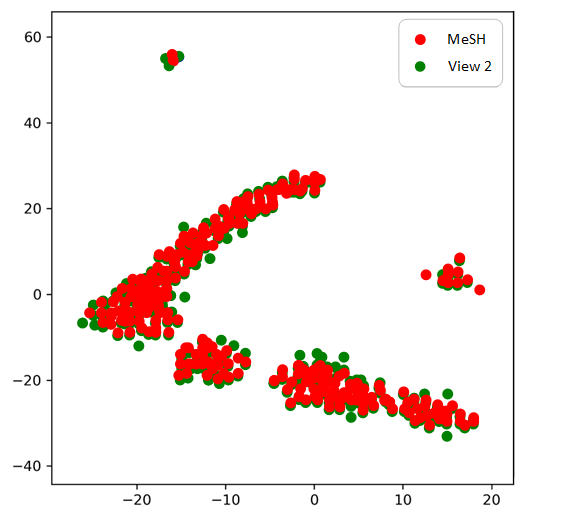} 
		\label{fig:imageB}
	}
	\caption{The t-SNE Visualization of Visual Features and MeSH in the MCA Module}
	\label{t-SNE}
\end{figure}

Fig. \ref{t-SNE} shows the t-SNE visualization of visual features aligned with MeSH information through GAN. Through visual analysis, we observed a high degree of overlap between the two sets of visual features and MeSH data in the low-dimensional space. This result indicates that MCA successfully aligns MeSH with visual features.

\subsubsection{Visualization Fine-Grained Alignment}
\begin{figure}[h]
	\centering
	\includegraphics[width=\columnwidth]{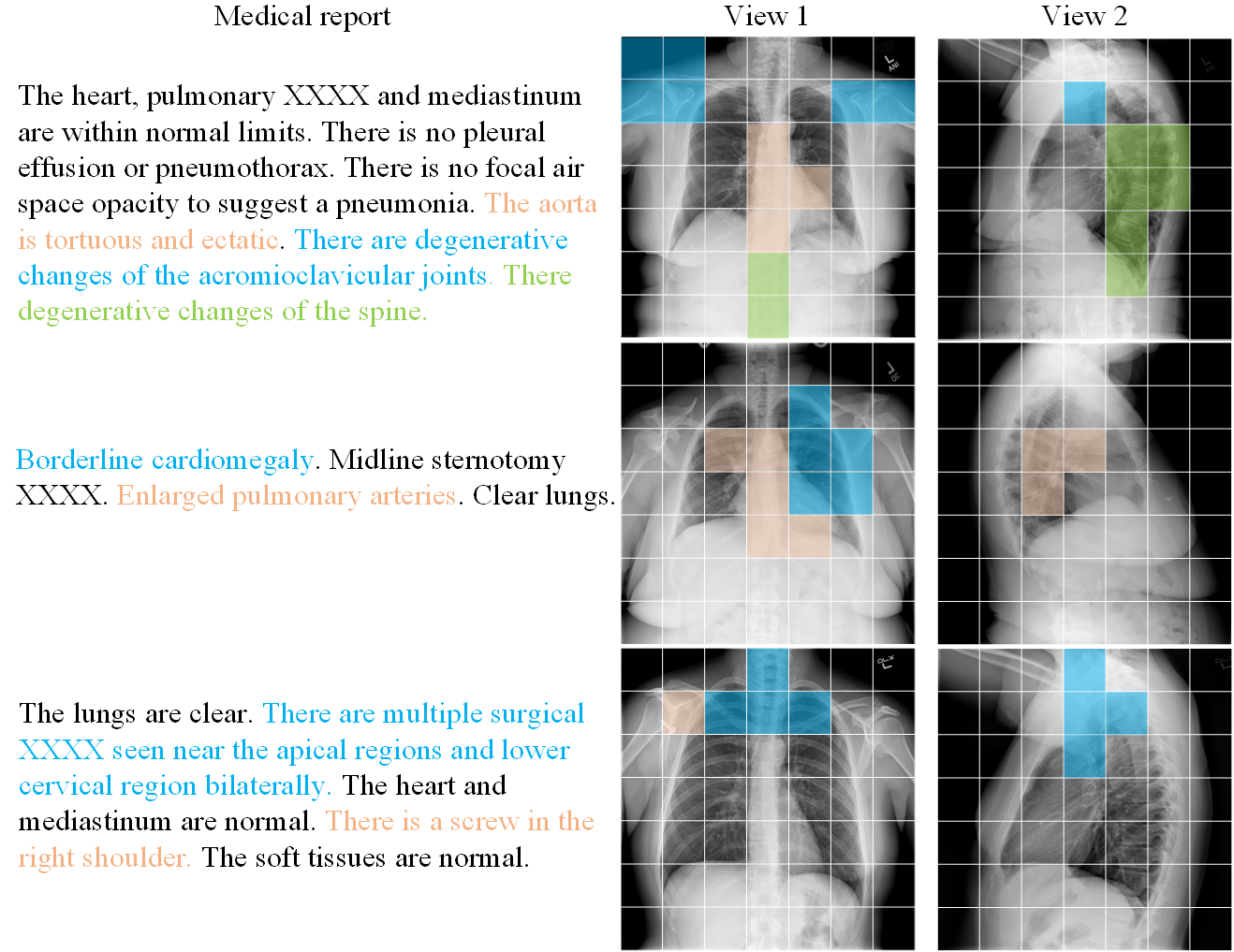}
	\caption{Visualization of alignment examples using hypergraph matching. Text phrases in the report and image regions, color-coded similarly, represent alignments established by hypergraph matching. Different colors distinguish multiple alignment relationships.}
	\label{align}
\end{figure}

Fig \ref{align} shows the alignment between visual regions and report phrases using color-coding, validating the effectiveness of hypergraph matching. For instance, the phrase "There is a screw in the right shoulder" aligns with the corresponding pink region in the image. However, some images from certain perspectives reveal incomplete correspondence between pathological phrases and their regions, highlighting the need to consider multi-view visual information to enhance alignment accuracy and comprehensiveness.

\subsection{Evaluation of Zero-Shot Performance}

We evaluated the zero-shot capabilities of DAMPER on the MIMIC-ABN dataset \cite{MIMIC-ABN}. Specifically, we trained the model on the MIMIC-CXR dataset and subsequently assessed its performance on the MIMIC-ABN test set. As shown in Table \ref{tab-2}, while DAMPER performed slightly below the ICON model on BLEU-1 and ROUGE-L metrics, it demonstrated superior results on BLEU-2 through BLEU-4 and METEOR metrics. These results highlight DAMPER's notable advantages in capturing semantic fluency and overlapping textual information. This performance underscores DAMPER's strong generalization capability in zero-shot medical report generation tasks, providing robust evidence of its potential for clinical applications.

\begin{table}[h]
	\centering
	\caption{Zero-shot test results of DAMPER on the MIMIC-ABN dataset.}
	\vspace{-10pt}
	\label{tab-2}
	\begin{center}
		\resizebox{\columnwidth}{!}{
			\begin{tabular}{c | c c c c c c}
				\toprule[1.25pt]
				Methods &B-1 &B-2 &B-3 &B-4 &M &R-L\\
				\hline
				R2Gen (EMNLP'20)&0.290 &0.157 &0.093 &0.061 &0.105 &0.208\\
				R2GenCMN (ACL'21)&0.264 &0.140 &0.085 &0.056 &0.098 &0.212\\
				ORGAN (ACL'23) &0.314 &0.180 &0.114 &0.078 &0.120 &0.234\\
				RECAP (EMNLP'23)&0.321 &0.182 &0.116 &0.080 &0.120 &0.223\\
				ICON (EMNLP'24)&\textbf{0.337} &\textbf{0.195} &0.126 &0.086 &0.129 &\textbf{0.236}\\
				\hline
				DAMPER &0.316 &\textbf{0.195} &\textbf{0.134} &\textbf{0.091} &\textbf{0.131} &0.209\\
				\bottomrule[1.25pt]
			\end{tabular}	
		}
	\end{center}
\end{table}

\section{Conclusion}
This study introduces DAMPER, a dual-stage medical report generation framework inspired by the clinical report-writing process. DAMPER consists of MCG and HFG stages to simulate physicians' overall observation and detailed examination, respectively. In MCG, the MCA module employs GAN to extract visual features corresponding to MeSH terms, which are input into the CMG to generate guiding MeSH. HFG constructs hypergraphs to model complex relationships between modality elements, designing Intra-RCA and Inter-RCA modules to generate fine-grained visual features aligning with detailed report descriptions. Finally, the outputs from MCA, CMG, and Intra-RCA are integrated into HFG's report decoder to produce accurate medical reports. Experiments conducted on the IU-Xray and MIMIC-CXR datasets demonstrate that DAMPER outperforms state-of-the-art methods. Ablation studies validate the effectiveness of each module, while visualization results highlight DAMPER’s ability to generate reports that accurately reflect pathological conditions. Furthermore, zero-shot testing results reveal DAMPER's strong generalization capability, as evidenced by its outstanding performance on the MIMIC-ABN test set.

\clearpage

\section{Acknowledgments}

This work was supported by the National Natural Science Foundation of China under Grant 82261138629 and 12326610;  Guangdong Provincial Key Laboratory under Grant 2023B1212060076, and Shenzhen Municipal Science and Technology Innovation Council under Grant JCYJ20220531101412030; and in part by the project of Shenzhen Science and Technology Innovation Committee under Grant JCYJ20240813141424032, the Foundation for Young innovative talents in ordinary universities of Guangdong  under Grant 2024KQNCX042.

\bibliography{refer}

\end{document}